# Large-Scale Music Annotation and Retrieval : Learning to Rank in Joint Semantic Spaces

Jason Weston, Samy Bengio, and Philippe Hamel

Google, USA
{jweston,bengio,hamelphi}@google.com

**Abstract.** Music prediction tasks range from predicting tags given a song or clip of audio, predicting the name of the artist, or predicting related songs given a song, clip, artist name or tag. That is, we are interested in every semantic relationship between the different musical concepts in our database. In realistically sized databases, the number of songs is measured in the hundreds of thousands or more, and the number of artists in the tens of thousands or more, providing a considerable challenge to standard machine learning techniques. In this work, we propose a method that scales to such datasets which attempts to capture the semantic similarities between the database items by modeling audio, artist names, and tags in a single low-dimensional semantic space. This choice of space is learnt by optimizing the set of prediction tasks of interest jointly using multi-task learning. Our method both outperforms baseline methods and, in comparison to them, is faster and consumes less memory. We then demonstrate how our method learns an interpretable model, where the semantic space captures well the similarities of interest.

## 1 Introduction

Users of software for annotating, retrieving and suggesting music are interested in a variety of tools that are all more or less related to the semantic interpretation of the audio, as perceived by the human listener. Such tasks include: (i) suggesting the next song to play given either one or many previously played songs, possibly with a set of ratings provided by the user, (ii) suggesting an artist to discover who is previously unknown to the user, given a set of rated artists, albums or songs (iii) browsing or searching by genre, style or mood. Several well known systems such as iTunes, `www.pandora.com` or `www.lastfm.com` are attempting to perform these tasks.

The audio itself for these tasks, in the form of songs, can easily be counted in the hundreds of thousands or more, and the number of artists in the tens of thousands or more in a large scale system. We might note that such data exhibits a typical "long tail" distribution where a small number of artists are very popular. For these artists one can collect lots of labeled data in the form of user plays, ratings and tags, while for the remaining large number of artists one has significantly less information (which we will refer to as "data sparsity"). At the extreme, users may have audio in their collection that was made by a local band or by themselves for which no other information is known (ratings, genres, or even the artist name). All one has in that case is the audio itself. Yet still, one may be interested in all the tasks described above with respect to these songs.

In this paper we describe a single unified model that can solve *all* the tasks described above in a large scale setting. The final model is lightweight in terms of memory usage, and provides reasonably fast test times, and hence could readily be used in a real system. The model we consider learns to represent audio, tags, and artist names jointly in a single low-dimensional embedding space. The low-dimension means our model has small capacity and we argue that this helps to deal with the problem of data sparsity. Simultaneously, the small number of parameters means that the memory usage is low.

To build a unified model, all of our tasks are trained jointly via multi-tasking, sharing the same embedding space, i.e. the same model parameters. In order to do that, we use a recently developed embedding algorithm [1], which was applied to a vision task, and extend it to perform multi-tasking (and apply it to the music annotation and retrieval domain). For each task, the parameters of the model that embed the entities of interest into the low dimensional space are learnt in order to optimize the criterion of interest, which is the precision at $k$ of the ranked list of retrieved entities. Typically, the tasks aim to learn that particular entities (e.g. audio and tags) should be close to each other in the embedding space. Hence, the distances in the embedding space can then be used for annotation or providing similar entities.

The model that we learn exhibits strong performance on all the tasks we tried, outperforming the baselines, and we also show that by multi-tasking all the tasks together the performance of our model improves. We argue that the reason for this improvement is that all of the tasks rely on the same semantic understanding of audio, artists and tags, and hence learning them together provides more information for each task. Finally, we show that the model indeed learns a rich semantic structure by visualizing the learnt embedding space. Semantically consistent entities appear close to each other in the embedding space.

The structure of the rest of the paper is as follows. Section 2 defines the tasks that we will consider. Section 3 describes the joint embedding model that we will employ, and Section 4 describes how to train (i.e., learn the parameters of) this model. Section 5 details prior related work, Section 6 describes our experiments, and Section 7 concludes.

## 2   Music Annotation and Retrieval Tasks

*Task Definitions:* In this work, we focus on being able to solve the following annotation and retrieval tasks:

1. **Artist prediction:** Given a song or audio clip (not seen at training time), return a ranked list of the likely artists to have performed it.
2. **Song prediction:** Given an artist's name, return a ranked list of songs (not seen at training time) that are likely to have been performed by that artist.

3. **Similar Artists:** Given an artist's name, return a ranked list of artists that are similar to that artist. Training data may or may not be provided for this task.
4. **Similar Songs:** Given a song or audio clip (not seen at training time), return a ranked list of songs that are similar to it.
5. **Tag prediction:** Given a song or audio clip (not seen at training time), return a ranked list of tags (e.g. rock, guitar, fast, ...) that might best describe the song.

*Evaluation:* In all cases, when a ranked list is returned we are interested in the correctness of the top of the ranked list, e.g. in the first $k \approx 15$ positions. For this reason, we measure the precision@$k$ for various small values of $k$:

$$\text{precision@}k = \frac{\text{number of true positives in the top } k \text{ positions}}{k}.$$

*Database:* We suppose we are given a database containing artist names, songs (in the form of features corresponding to their audio content), and tags. We will denote our training data as triplets of the following form:

$$\mathcal{D} = \{(a_i, t_i, s_i)\}_{i=1,\ldots,m} \in \{1,\ldots,|\mathcal{A}|\}^{|a_i|} \times \{1,\ldots,|\mathcal{T}|\}^{|t_i|} \times \mathbb{R}^{|\mathcal{S}|},$$

where each triplet represents a song indexed by $i$: $a_i$ are the artist features, $t_i$ are the tag features and $s_i$ are the audio (sound) features.

Each song has attributed to it a set of artists $a_i$, where each artist is indexed from 1 to $|\mathcal{A}|$ (indices into a dictionary of artist names). Hence, a given song can have multiple artists, although it usually only has one and hence $|a_i| = 1$. Similarly, each song may also have a corresponding set of tags $t_i$, where each tag is indexed from 1 to $|\mathcal{T}|$ (indices into a dictionary of tags).

The audio of the song itself is represented as an $|\mathcal{S}|$-dimensional real-valued feature vector $s_i$. In this work we do not focus on developing novel feature representations for audio (instead, we will develop learning algorithms that use these features). Hence, we will use standard feature representations that can be found in the literature. More details on the features we use to represent audio are given in Section 6.2.

## 3 Semantic Embedding Model for Music Understanding

The core idea in our model is that songs, artists and tags attributed to music can all be reasoned about jointly by learning a single model to capture the semantics of, and hence the relationships between, each of these musical concepts.

Our method makes the assumption that these semantic relationships can be modeled in a feature space of dimension $d$, where musical concepts (songs, artists or tags) are represented as coordinate vectors. The similarity between two concepts is measured using the dot product between their two vector representations. The vectors will be learnt to induce similarities relevant (i.e. optimize the precision@k metric) for the tasks defined in Section 2.

For a given artist, indexed by $j \in 1, \ldots, |\mathcal{A}|$, its coordinate vector is expressed as:
$$\Phi_{Artist}(i) : \{1, \ldots, |\mathcal{A}|\} \to \mathbb{R}^d = A_i.$$
where $A = [A_1, \ldots, A_{|\mathcal{A}|}]$ is a $d \times |\mathcal{A}|$ matrix of the parameters (vectors) of all the artists in the database. This entire matrix will be learnt during the learning phase of the algorithm.

Similarly, for a given tag, indexed by $j \in 1, \ldots, |\mathcal{T}|$, its coordinate vector is expressed as:
$$\Phi_{Tag}(i) : \{1, \ldots, |\mathcal{T}|\} \to \mathbb{R}^d = T_i.$$
where $T = [T_1, \ldots, T_{|\mathcal{T}|}]$ is a $d \times |\mathcal{T}|$ matrix of the parameters (vectors) of all the tags in the database. Again, this entire matrix will also be learnt during the learning phase of the algorithm.

Finally, for a given song or audio clip we consider the following function that maps its audio features $s'$ to a $d$-dimensional vector using a linear transform $V$:
$$\Phi_{Song}(s') : \mathbb{R}^{|\mathcal{S}|} \to \mathbb{R}^d = Vs'.$$
The $d \times |\mathcal{S}|$ matrix $V$ will also be learnt.

We also choose for our family of models to have constrained norm:
$$||A_i||_2 \leq C, \quad i = 1, \ldots, |\mathcal{A}|, \tag{1}$$
$$||T_i||_2 \leq C, \quad i = 1, \ldots, |\mathcal{T}|, \tag{2}$$
$$||V_i||_2 \leq C, \quad i = 1, \ldots, |\mathcal{S}|, \tag{3}$$
using the hyperparameter $C$ which will act as a regularizer in a similar way as used in lasso [2].

Our overall goal is, for a given input, to rank the possible outputs of interest depending on the task (see Section 2 for the list of tasks) such that the highest ranked outputs are the best semantic match for the input. For example, for the artist prediction task, we consider the following ranking function:
$$f_i^{ArtistPred}(s') = f_i^{AP}(s') = \Phi_{Artist}(i)^\top \Phi_{Song}(s') = A_i^\top V s' \tag{4}$$
where the possible artists $i \in \{1, \ldots, |\mathcal{A}|\}$ are ranked according to the magnitude of $f_i(x)$, largest first. Similarly, for song prediction, similar artists, similar songs and tag prediction we have the following ranking functions:
$$f_{s'}^{SongPred}(i) = f_{s'}^{SP}(i) = \Phi_{Song}(s')^\top \Phi_{Artist}(i) = (Vs')^\top A_i \tag{5}$$
$$f_j^{SimArtist}(i) = f_j^{SA}(i) = \Phi_{Artist}(j)^\top \Phi_{Artist}(i) = A_j^\top A_i \tag{6}$$
$$f_{s'}^{SimSong}(s'') = f_{s'}^{SS}(s'') = \Phi_{Song}(s')^\top \Phi_{Song}(s'') = (Vs')^\top Vs'' \tag{7}$$
$$f_i^{TagPred}(s') = f_i^{TP}(s') = \Phi_{Tag}(i)^\top \Phi_{Song}(s') = T_i^\top V s'. \tag{8}$$

Note that many of these tasks *share* the same parameters, for example the song prediction and similar artist tasks share the matrix $A$ whereas the tag prediction and song prediction tasks share the matrix $V$. As we shall see, it is possible to learn the parameters $A$, $T$ and $V$ of our model jointly to perform well on all our tasks, which is referred to as multi-task learning [3]. In the next section we describe how we train our model.

## 4  Training the Semantic Embedding Model

During training, our objective is to learn the parameters of our model that provide good ranking performance on the training set, using the precision at $k$ measure (with the overall goal that this also generalizes to performing well on our test data, of course). We want to achieve this simultaneously for all the tasks at once using multi-task learning.

### 4.1  Multi-Task Training

Let us suppose we define the objective function for a given task as $\sum_i err(f(x_i), y_i)$ where $x$ is the set of input examples, and $y$ are the set of targets for these examples, and $err$ is a loss function that measures the quality of a given ranking (the exact form of this function will be discussed in Section 4.2).

In the case of the tag prediction task we wish to minimize the function $\sum_i err(f^{TP}(s_i), t_i)$ and for the artist prediction task we wish to minimize the function $\sum_i err(f^{AP}(s_i), a_i)$. To multi-task these two tasks we simply consider the (unweighted) sum of the two objectives:

$$err^{AP+TP}(\mathcal{D}) = \sum_{i=1}^m err(f^{AP}(s_i), a_i) + \sum_{i=1}^m err(f^{TP}(s_i), t_i).$$

We will optimize this function by stochastic gradient descent [4]. This amounts to iteratively repeating the following procedure [3]:

1. Pick one of the tasks at random.
2. Pick one of the training input-output pairs for this task.
3. Make a gradient step for this task and input-output pair.

The procedure is the same when considering more than two tasks.

### 4.2  Loss Functions

We consider two loss functions, the standard margin ranking loss and the newly introduced WARP (Weighted Approximately Ranked Pairwise) Loss [1].

**AUC Margin Ranking Loss** A standard loss function that is often using for retrieval is the margin ranking criterion [5,6], in particular it was used for text embedding models in [7]. Assuming the input $x$ and output $y$ (which can be replaced by artists, songs or tags, depending on the task) the loss is:

$$err_{AUC}(\mathcal{D}) = \sum_{i=1}^m \sum_{j \in y_i} \sum_{k \notin y_i} \max(0, 1 + f_k(x_i) - f_j(x_i)) \qquad (9)$$

which considers all pairs of positive and negative labels, and assigns each a cost if the negative label is larger or within a "margin" of 1 from the positive label. Optimizing this loss is similar to optimizing the area under the curve of the receiver operating characteristic curve. That is, all pairwise violations are considered equally if they have the same margin violation, independent of their position in the list. For this reason the margin ranking loss might not optimize precision at $k$ very accurately.

**WARP Loss** To focus more on the top of the ranked list, where the top $k$ positions are those we care about using the precision at $k$ measure, one can *weigh* the pairwise violations depending on their position in the ranked list. This type of ranking error functions was recently developed in [8], and then used in an image annotation application in [1]. These works consider a class of ranking error functions:

$$err_{WARP}(\mathcal{D}) = \sum_{i=1}^{m} \sum_{j \in y_i} L(rank_j^1(f(x_i))) \tag{10}$$

where $rank_j^1(f(x_i))$ is the margin-based rank of the true label $j \in y_i$ given by $f(x_i)$:

$$rank_j^1(f(x_i)) = \sum_{k \notin y_i} I(1 + f_k(x_i) \geq f_j(x_i))$$

where $I$ is the indicator function, and $L(\cdot)$ transforms this rank into a loss:

$$L(r) = \sum_{i=1}^{r} \alpha_i, \text{ with } \alpha_1 \geq \alpha_2 \geq \cdots \geq 0. \tag{11}$$

Different choices of $\alpha$ define different weights (importance) of the relative position of the positive examples in the ranked list. In particular:

- For $\alpha_i = 1$ for all $i$ we have the same AUC optimization as equation (9).
- For $\alpha_1 = 1$ and $\alpha_{i>1} = 0$ the precision at 1 is optimized.
- For $\alpha_{i \leq k} = 1$ and $\alpha_{i \geq k} = 0$ the precision at $k$ is optimized.
- For $\alpha_i = 1/i$ a smooth weighting over positions is given, where most weight is given to the top position, with rapidly decaying weight for lower positions. This is useful when one wants to optimize precision at $k$ for a variety of different values of $k$ at once [8].

We will optimize this function by stochastic gradient descent following the authors of [1], that is samples are drawn at random, and a gradient step is made for that sample. As in that work, due to the cost of computing the exact rank in (10) it is approximated by sampling. That is, for a given positive label, one draws negative labels until a violating pair is found, and then approximates the rank with[1]

$$rank_j^1(f(x_i)) \approx \left\lfloor \frac{Y-1}{N} \right\rfloor$$

---

[1] In fact, this gives a biased estimator of the rank, but as we are free to choose the vector $\alpha$ in any case one could imagine correcting it by slightly adjusting the weights. In fact, the sampling process gives an unbiased estimator if we consider a

**Algorithm 1** MUSLSE training algorithm.
---
**Input:** labeled data for several tasks.
Initialize model parameters (we use mean 0, standard deviation $\frac{1}{\sqrt{d}}$).
**repeat**
  Pick a random task, and let $f(x') = \Phi_{Output}(y')^\top \Phi_{Input}(x')$ be the prediction function for that task, and let $x$ and $y$ be its input and output examples, where there are $Y$ possible output labels.
  Pick a random labeled example $(x_i, y_i)$ (for the task chosen).
  Pick a random positive label $j \in y_i$ for $x_i$.
  Compute $f_j(x_i) = \Phi_{Output}(j)^\top \Phi_{Input}(x_i)$
  Set $N = 0$.
  **repeat**
    Pick a random negative label $k \in \{1, \ldots, Y\} \notin y_i$.
    Compute $f_k(x_i) = \Phi_{Output}(k)^\top \Phi_{Input}(x_i)$
    $N = N + 1$.
  **until** $f_k(x_i) > f_j(x_i) - 1$ or $N \geq Y - 1$
  **if** $f_k(x_i) > f_j(x_i) - 1$ **then**
    Make a gradient step to minimize:
       $L(\lfloor \frac{Y-1}{N} \rfloor)|1 - f_j(x_i) + f_k(x_i)|_+$
    Project weights to enforce constraints (1)-(3).
  **end if**
**until** validation error does not improve.
---

where $\lfloor . \rfloor$ is the floor function, $Y$ is the number of output labels (which is task dependent, e.g. $Y = |\mathcal{T}|$ for the tag prediction task) and $N$ is the number of trials in the sampling step. Intuitively, if we need to sample more negative labels before we find a violator, then the rank of the true label is likely to be small (it is likely to be at the top of the list, as few negatives are above it).

Pseudocode of training our method which we call MUSLSE (Music Understanding by Semantic Large Scale Embedding, pronounced "muscles") using the WARP loss is given in Algorithm 1. We use a fixed learning rate $\gamma$, chosen using a validation set (a decaying schedule over time $t$ is also possible, but we did not implement that approach). The validation error in the last line of Algorithm 1 is in practice evaluated every so often for computational efficiency.

**Training Ensembles** In our experiments, we will use the training schemes just described above for models of dimension $d = 100$. To train models with larger dimension we build an ensemble of several MUSLSE models. That is, for dimension $d = 300$ we would train three models. As we use stochastic gradient descent, each of the models will learn slightly different model parameters. When

---

new function $\tilde{L}$ instead of $L$ in Equation (10), with:

$$\tilde{L}(k) = E\left[L\left(\left\lfloor \frac{Y-1}{N_k} \right\rfloor\right)\right].$$

Hence, this approach defines a slightly different ranking error.

averaging their ranking scores, $f_i^{ensemble}(x) = f_i^1(x) + f_i^2(x) + f_i^3(x)$ for a given label $i$ one can obtain improved results, as has been shown in [1] on vision tasks.

## 5   Related Approaches

The task of automatically annotating music consists of assigning relevant tags to a given audio clip. Tags can represent a wide range of concepts such as genre (rock, pop, jazz, etc.), instrumentation (guitar, violon, etc.), mood (sad, calm, dark, etc.), locale (Seattle, NYC, Indian), opinions (good, love, favorite) or any other general attribute of the music (fast, eastern, wierd, etc.). A set of tags gives us a high-level semantic representation of a clip than can the be useful for other tasks such as music recommendation, playlist generation or music similarity measure. Most automatic annotation systems are built around the following recipe. First, features are extracted from the audio. These features often include MFCCs (section 6.2) and other spectral or temporal features. The features can also be learnt directly from the audio [9]. Then, these features are aggregated or summarized over windows of a given length, or over the whole clip. Finally, some machine learning algorithm is trained over these features in order to obtain a classifier for each tag. Often, the machine learning algorithm attempts to model the semantic relations between the tags [10]. A few state-of-the-art automatic annotation systems are briefly described in section 6.3. A more extensive review of the automatic tagging of audio is presented in [11].

Artist and song similarity is at the core of most music recommendation or playlist generation systems. However, music similarity measures are subjective, which makes it difficult to rely on ground truth. This makes the evaluation of such systems more complex. This issue is addressed in [12] and [13].

These tasks can be tackled using content-based features or meta-data from human sources. Features commonly used to predict music similarity include audio features, tags and collaborative filtering information.

Meta-data such as tags and collaborative filtering data have the advantage of considering human perception and opinions. These concepts are important to consider when building a music similarity space. However, meta-data suffers from a popularity bias, because a lot of data is available for popular music, but very little information can be found on new or less known artists. In consequence, in systems that rely solely upon meta-data, everything tends to be similar to popular artists. Another problem, known as the cold-start problem, arises with new artists or songs for which no human annotation exists yet. It is then impossible to get a reliable similarity measure, and is thus difficult to correctly recommend new or less known artists.

Content-based features such as MFCCs, spectral features and temporal features have the advantage of being easily accessible, given the audio, and do not suffer from the popularity bias. However, audio features cannot take into account the social aspect of music. Despite this, a number of music similarity systems rely only on acoustic features [14, 15].

Ideally, we would like to integrate those complementary sources of information in order to improve the performance of the similarity measure. Several systems such as [16, 17] combine audio content with meta-data. One way to do this is to embed songs or artists in a Euclidean space using metric learning [18].

We should also note that other related work (outside of the music domain) includes learning embeddings for supervised document ranking [7], semi-supervised multi-task learning [19, 20] and for vision tasks [21, 1].

## 6 Experiments

### 6.1 Datasets

**TagATune Dataset** The TagATune dataset consists of a set of 30 second clips with annotations. Each clip is annotated with one or more descriptors, or tags, that represent concepts that can be associated with the given clip. The set of descriptors also include negative concepts (no voice, not classical, no drums, etc.). The annotations of the dataset were collected with the help of a web-based game. Details of how the data was collected are described in [22].

The TagATune dataset was used in the MIREX 2009 contest on audio tag classification [23]. In order to be able to compare our results with the MIREX 2009 contestants, we used the same set of tags and the same train/test split as in the contest.

**Big-data Dataset** We had access to a large proprietary database of tracks and artists, from which we took a subset for this experimental study.

We processed this data similarly to TagATune. In this case we only considered using MFCC features (see Section 6.2). We evaluate the artist prediction, song prediction and song similarity tasks on this dataset. The test set (which is the same test set for all tasks) contains songs not previously seen in the training set.

As mentioned in section 5, it is difficult to obtain reliable ground truth for music similarity tasks. In our experiments, song similarity is evaluated by taking all songs by the same artist as a given query song as positives, and all other songs as negatives. We do not evaluate the similar artist task due to not having labeled data, however our model would be perfectly capable of working on this type of data as well.

Table 1 provides summary statistics of the number of songs and labels for the TagATune and Big-data datasets used in our experiments.

### 6.2 Audio Feature Representation

In this work we focus on learning algorithms, not feature representations. We used the well-known Mel Frequency Cepstral Coefficient (MFCC) representation. MFCCs take advantage of source/filter deconvolution from the cepstral transform and perceptually-realistic compression of spectra from the Mel pitch scale. They have been used extensively in the speech recognition community for many years [24] and are also the de facto baseline feature used in music modeling (see for instance [25]). In particular, the MFCCs are known to offer a reasonable representation of the musical timbre [26]. In this paper, 13 MFCCs were extracted every 10ms over a hamming window of 25ms, and first and second derivatives were concatenated, for a total of 39 features. We then computed a dictionary of $d = 2000$ typical MFCC vectors over the training set (using K-means) and represented each song as a vector of counts, over the set of frames in the given song, of the number of times each dictionary vector was nearest to the frame in the MFCC space. The resulting feature vectors thus have dimension $d = 2000$ with an average of $|\mathcal{S}|_{\bar{\varnothing}} = 1032$ non-zero values. It takes on average 2 seconds to extract these features per song.

Table 1. Summary statistics of the datasets used in this paper.

| Statistics | TagATune | Big-data |
|---|---|---|
| Number of Training+Validation Songs/Clips | 16,289 | 275,930 |
| Number of Test Songs | 6,499 | 66,072 |
| Number of Tag Labels | 160 | - |
| Number of Artist Labels | - | 26,972 |

Our second set of features, Stabilized Auditory Image (SAI) features are based on adaptive pole-zero filter cascade (PZFC) auditory filterbanks, followed by a sparse coding step similar to the one used for our MFCC features. They have been used successfully in audio retrieval tasks [27]. Our implementation yields a sparse representation of $d = 7168$ features with an average of $|\mathcal{S}|_{\bar{\varnothing}} = 4000$ non-zero values. It takes on average 6 seconds to extract these features per song. In our experiments, we consider using either MFCC features, or we use jointly the two sets of features by concatenating their respective vector representation (MFCC+SAI).

### 6.3  Baselines

We compare our proposed approach to the following baselines: one-versus-rest large margin classifiers (one-vs-rest) of the form $f_i(x) = w_i^\top x$ trained using the margin perceptron algorithm, which gives similar results to support vector machines [28]. The loss function for tag prediction in that case is:

$$\sum_{i=1}^{m} \sum_{j=1}^{|\mathcal{T}|} \max(0, 1 - \phi(t_i, j) f_i(a_i))$$

where $\phi(t', j) = 1$ if $j \in t'$, and $-1$ otherwise.

For the similar song task we compare to using cosine similarity in the feature space, a classical information retrieval baseline [29].

Additionally, on the TagATune dataset we compare to all the entrants of the MIREX 2009 competition [23]. The performance of the different models are described in detail at http://www.music-ir.org/mirex/wiki/2009:Audio_Tag_

`Classification_Tagatune_Results`. All the algorithms in the competition follow more or less the same general pattern described in Section 5. We present here the results of the four best contestants: Marsyas [30], Mandel [31], Manzagol [32] and Zhi [33]. Every submission uses MFCCs as features, except for Mandel, which computes another kind of cepstral transform, quite similar to MFCCs. Furthermore, Mandel also uses a set of temporal features and Marsyas adds a set of spectral features: spectral centroid, rolloff and flux. All the submissions use a temporal aggregation of the features, though the methods used vary. The classification algorithms also varied.

The Marsyas algorithm uses running means and standard deviations of the features as input to a two-stage SVM classifier. The second stage SVM helps to capture the relations between tags. The Mandel submission uses balanced SVMs for each tag. In order to balance the training set for a given tag, a number equal to the number of positive examples is chosen at random in the non-positive examples to form the training set for that given tag. Manzagol uses vector quantization and applies an algorithm called PAMIR (passive-aggressive model for image retrieval) [5]. Finally, Zhi also uses Gaussian Mixture Models to obtain a song-level representation and uses a semantic multiclass labeling model.

### 6.4  Results

**TagATune Results**  The results of comparing all the methods on the tag prediction task on the TagATune data are summarized in Table 2. MUSLSE outperforms the one-vs-rest baseline that we ran using the same features, as well as the competition entrants on the TagATune dataset. Results of choosing different embedding dimensions $d$ for MUSLSE are given in Table 5 and show that the performance is relatively stable over different choices of $d$, although we see slight improvements for larger $d$. We give a more detailed analysis of the results, including time and space requirements in subsequent sections.

Table 2. **Summary of Test Set Results on TagATune.** Precision at 3, 6, 9, 12 and 15 are given. Our approach, MUSLSE, with embedding dimension $d = 400$, outperforms the baselines.

| Algorithm | Features | p@3 | p@6 | p@9 | p@12 | p@15 |
|---|---|---|---|---|---|---|
| Zhi | MFCC | 0.224 | 0.192 | 0.168 | 0.146 | 0.127 |
| Manzagol | MFCC | 0.255 | 0.194 | 0.159 | 0.136 | 0.119 |
| Mandel | cepstral + temporal | 0.323 | 0.245 | 0.197 | 0.167 | 0.145 |
| Marsyas | spectral features + MFCC | 0.440 | 0.314 | 0.244 | 0.201 | 0.172 |
| one-vs-rest | MFCC | 0.349 | 0.244 | 0.193 | 0.154 | 0.136 |
| MUSLSE | MFCC | 0.382 | 0.275 | 0.219 | 0.182 | 0.157 |
| one-vs-rest | MFCC+SAI | 0.362 | 0.261 | 0.221 | 0.167 | 0.151 |
| MUSLSE | MFCC+SAI | 0.473 | 0.330 | 0.256 | 0.211 | 0.179 |

**Table 3. WARP vs. AUC optimization.** Precision at $k$ for various values of $k$ training with AUC or WARP loss using MUSLSE on the TagATune dataset. WARP loss improves over AUC.

| Algorithm | Loss | Features | d | p@3 | p@6 | p@9 | p@12 | p@15 |
|---|---|---|---|---|---|---|---|---|
| MUSLSE | AUC | MFCC | 100 | 0.226 | 0.179 | 0.147 | 0.128 | 0.112 |
| MUSLSE | WARP | MFCC | 100 | 0.371 | 0.267 | 0.212 | 0.177 | 0.153 |
| MUSLSE | AUC | MFCC | 400 | 0.222 | 0.179 | 0.151 | 0.131 | 0.116 |
| MUSLSE | WARP | MFCC | 400 | 0.382 | 0.275 | 0.219 | 0.182 | 0.157 |
| MUSLSE | AUC | MFCC + SAI | 100 | 0.301 | 0.217 | 0.175 | 0.147 | 0.128 |
| MUSLSE | WARP | MFCC + SAI | 100 | 0.452 | 0.319 | 0.248 | 0.205 | 0.174 |
| MUSLSE | AUC | MFCC + SAI | 400 | 0.338 | 0.248 | 0.199 | 0.166 | 0.143 |
| MUSLSE | WARP | MFCC + SAI | 400 | 0.473 | 0.33 | 0.256 | 0.211 | 0.179 |

**AUC via WARP loss** We compared MUSLSE embedding models trained with either WARP or AUC optimization for different embedding dimensions and feature types. The results given in Table 3 show WARP gives superior precision @ k for all the parameters tried.

**Tag Embeddings on TagATune** Example tag embeddings learnt by MUSLSE for the TagATune data are given in Table 4. We observe that the embeddings capture the semantic structure of the tags (and note that songs are also embedded in this same space).

**Table 4. Related tags in the embedding space** learnt by MUSLSE ($d = 400$, using MFCC+SAI features) on the TagATune data. We show the closest five tags (from the set of 160 tags) in the embedding space using the similarity measure $\Phi_{Tag}(i)^\top \Phi_{Tag}(j) = T_i^\top T_j$.

| Tag | Neighboring Tags |
|---|---|
| female opera | opera, operatic, woman, male opera, female singer |
| hip hop | rap, talking, funky, punk, funk |
| middle eastern | eastern, sitar, indian, oriental, india |
| flute | flutes, wind, clarinet, oboe, horn |
| techno | electronic, dance, synth, electro, trance |
| ambient | new age, spacey, synth, electronic, slow |
| celtic | irish, fiddle, folk, medieval, female singer |

**Multi-Tasking Results on Big-data** Results comparing MUSLSE with the one-vs-rest and cosine similarity baselines for Big-data are given in Table 6. All methods use MFCC features, and MUSLSE uses $d = 100$. Two flavors of MUSLSE are presented: training on one of the tasks alone, or all three tasks jointly. The results show that MUSLSE performs well compared to the baseline approaches and that multi-tasking improves performance on all the tasks compared to training on a single task.

**Table 5. Changing the Embedding Size on TagATune.** Test Error metrics when we change the dimension $d$ of the embedding space used in MUSLSE for MFCC and MFCC+SAI features on the TagATune dataset.

| Algorithm | Features | p@3 | p@6 | p@9 | p@12 | p@15 |
|---|---|---|---|---|---|---|
| MUSLSE ($d = 100$) | MFCC | 0.371 | 0.267 | 0.212 | 0.177 | 0.153 |
| MUSLSE ($d = 200$) | MFCC | 0.379 | 0.273 | 0.216 | 0.180 | 0.156 |
| MUSLSE ($d = 300$) | MFCC | 0.381 | 0.273 | 0.217 | 0.181 | 0.157 |
| MUSLSE ($d = 400$) | MFCC | 0.382 | 0.275 | 0.219 | 0.182 | 0.157 |
| MUSLSE ($d = 100$) | MFCC+SAI | 0.452 | 0.319 | 0.248 | 0.205 | 0.174 |
| MUSLSE ($d = 200$) | MFCC+SAI | 0.465 | 0.325 | 0.252 | 0.208 | 0.177 |
| MUSLSE ($d = 300$) | MFCC+SAI | 0.470 | 0.329 | 0.255 | 0.209 | 0.178 |
| MUSLSE ($d = 400$) | MFCC+SAI | 0.473 | 0.33 | 0.256 | 0.211 | 0.179 |
| MUSLSE ($d = 600$) | MFCC+SAI | 0.477 | 0.334 | 0.259 | 0.212 | 0.180 |
| MUSLSE ($d = 800$) | MFCC+SAI | 0.476 | 0.334 | 0.259 | 0.212 | 0.181 |

**Table 6. Summary of Test Set Results on Big-data.** Precision at 1 and 6 are given for three different tasks. Our approach, MUSLSE outperforms the baseline approaches when training for an individual task, and provides improved performance when multi-tasking all tasks at once.

| Algorithm | Artist Prediction | | Song Prediction | | Similar Songs | |
|---|---|---|---|---|---|---|
| | p@1 | p@6 | p@1 | p@6 | p@1 | p@6 |
| one-vs-rest$^{ArtistPrediction}$ | 0.0551 | 0.0206 | - | - | - | - |
| cosine similarity | - | - | - | - | 0.0427 | 0.0159 |
| MUSLSE$^{SingleTask}$ | 0.0958 | 0.0328 | 0.0837 | 0.0406 | 0.0533 | 0.0225 |
| MUSLSE$^{AllTasks}$ | 0.1110 | 0.0352 | 0.0940 | 0.0433 | 0.0557 | 0.0226 |

**Computational Expense** A summary of the test time and space complexity of one-vs-rest compared to MUSLSE is given in Table 7 (not including cost of feature computation, see Section 6.2) as well as concrete numbers on our particular datasets using a single computer, and assuming the data fits in memory. One-vs-rest artist prediction takes around 2 seconds per song on the Big-data and requires 1.85 GB of memory. In contrast MUSLSE takes 0.045 seconds, and requires far less memory, only 27.7 MB. MUSLSE can be feasibly run on a laptop using limited resources whereas the memory requirements of one-vs-rest are rather high (and will be worse for larger database sizes). MUSLSE has a second advantage that it is not much slower at test time if we choose a larger and denser set of features, as it maps these features into a low dimensional embedding space and the bulk of the computation is then in that space.

**Table 7. Algorithm Time and Space Complexity.** Time and space complexity needed to return the top ranked tag on TagATune (or artist in Big-data) on a single test set song, not including feature generation using MFCC+SAI features. Prediction times (s=seconds) and memory requirements are also given, we report results for MUSLSE with $d = 100$. We denote by $Y$ the number of labels (tags or artists), $|\mathcal{S}|$ the music input dimension, $|\mathcal{S}|_{\bar{\varnothing}}$ the average number of non-zero feature values per song, and $d$ the size of the embedding space.

| Algorithm | Time Complexity | Space Complexity | Test Time and Memory Usage | | | |
|---|---|---|---|---|---|---|
| | | | TagATune | | Big-data | |
| | | | Time | Space | Time | Space |
| one-vs-rest | $\mathcal{O}(Y \cdot \|\mathcal{S}\|_{\bar{\varnothing}})$ | $\mathcal{O}(Y \cdot \|\mathcal{S}\|)$ | 0.012 s | 11.3 MB | 2.007 s | 1.85 GB |
| MUSLSE | $\mathcal{O}((Y + \|\mathcal{S}\|_{\bar{\varnothing}}) \cdot d)$ | $\mathcal{O}((Y + \|\mathcal{S}\|) \cdot d)$ | 0.006 s | 7.2 MB | 0.045 s | 27.7 MB |

## 7   Conclusions

We have introduced a music annotation and retrieval model that works by jointly learning several tasks by mapping entities of various types (audio, artist names and tags) into a single low-dimensional space where they all live. This appears to give a number of benefits, specifically:

(i) semantic similarities between all the entity types are learnt in the embedding space,
(ii) by multi-tasking all the tasks sharing the same embedding space we do have data for, accuracy improves for all tasks,
(iii) optimizing (approximately) the precision at $k$ leads to improved performance,
(iv) as the model has low-capacity this makes it harder to overfit on the tail of the distribution (where data is sparse),
(v) the model is also fast at test time and has low memory usage.

Our resulting model performed well compared to baselines on two datasets, and is scalable enough to use in a real-world system.

## 8   Acknowledgements

We thank Doug Eck, Ryan Rifkin and Tom Walters for providing us with the Big-data set and extracting the relevant features on it.